\def\year{2022}\relax
\documentclass[letterpaper]{article} 
\usepackage{aaai22}  
\usepackage{times}  
\usepackage{helvet}  
\usepackage{courier}  
\usepackage[hyphens]{url}  
\usepackage{graphicx} 
\usepackage{natbib}  
\usepackage{caption} 
\DeclareCaptionStyle{ruled}{labelfont=normalfont,labelsep=colon,strut=off} 
\usepackage{algorithm}
\usepackage{algorithmic}

\usepackage{amsmath}
\usepackage{amsfonts}
\usepackage{bm}

\def\eqref#1{equation~\ref{#1}}

\def\1{\bm{1}}

\DeclareMathAlphabet{\mathsfit}{\encodingdefault}{\sfdefault}{m}{sl}
\SetMathAlphabet{\mathsfit}{bold}{\encodingdefault}{\sfdefault}{bx}{n}

\newcommand{\R}{\mathbb{R}}

\DeclareMathOperator*{\argmax}{arg\,max}
\DeclareMathOperator*{\argmin}{arg\,min}

\usepackage{booktabs}       
\usepackage{amsfonts}       
\usepackage{nicefrac}       
\usepackage{microtype}      
\usepackage{times}
\usepackage{soul}
\usepackage{url}
\usepackage{amsmath}
\usepackage{amsthm}
\usepackage{bbm}
\usepackage{thmtools}
\usepackage{cleveref}
\usepackage{subcaption}     
\usepackage{multirow}
\usepackage{adjustbox}
\usepackage{mathtools}
\newcommand{\defeq}{\coloneqq}

\renewcommand{\r}{\mathbb{R}}

\newcommand{\y}{\mathcal{Y}}

\renewcommand{\r}[1]{\mathop{Pr}\limits_{#1}}

\def\i{{\mathbb{I}}}

\def\R{{\mathbb{R}}}

\newtheorem{definition}{Definition}
\newtheorem{proposition}{Proposition}
\newtheorem{remark}{Remark}
\usepackage{newfloat}
\usepackage{listings}
\floatstyle{ruled}
\newfloat{listing}{tb}{lst}{}
\floatname{listing}{Listing}
\title{Input-Specific Robustness Certification for Randomized Smoothing
}
\author {
    Ruoxin Chen\textsuperscript{\rm 1},
    Jie Li\textsuperscript{\rm 1} \thanks{Jie Li is the corresponding author, who is with the Department of Computer Science and Engineering and MoE Key Lab of Artificial Intelligence, AI Institute. Shanghai Jiao Tong University Shanghai, China. Email: lijiecs@sjtu.edu.cn},
    Junchi Yan\textsuperscript{\rm 1},
    Ping Li\textsuperscript{\rm 2},
    Bin Sheng\textsuperscript{\rm 1}\\
}

\affiliations {
    \textsuperscript{\rm 1} Shanghai Jiao Tong University \\
    \textsuperscript{\rm 2} 	The Hong Kong Polytechnic University \
}

\usepackage{bibentry}

\begin{document}

\maketitle
\begin{abstract}
 Although randomized smoothing has demonstrated high certified robustness and superior scalability to other certified defenses, the high computational overhead of the robustness certification bottlenecks the practical applicability, as it depends heavily on the large sample approximation for estimating the confidence interval. In existing works, the sample size for the confidence interval is universally set and agnostic to the input for prediction. This Input-Agnostic Sampling (IAS) scheme may yield a poor Average Certified Radius (ACR)-runtime trade-off which calls for improvement. In this paper, we propose Input-Specific Sampling (ISS) acceleration to achieve the cost-effectiveness for robustness certification, in an adaptive way of reducing the sampling size based on the input characteristic. Furthermore, our method universally controls the certified radius decline from the ISS sample size reduction. The empirical results on CIFAR-10 and ImageNet show that ISS can speed up the certification by more than three times at a limited cost of 0.05 certified radius. Meanwhile, ISS surpasses IAS on the average certified radius across the extensive hyperparameter settings. Specifically, ISS achieves ACR=0.958 on ImageNet ($\sigma=1.0$) in 250 minutes, compared to ACR=0.917 by IAS under the same condition. We release our code in \url{https://github.com/roy-ch/Input-Specific-Certification}.
\end{abstract}
\section{Introduction}
\label{sec:intro}
Neural networks are known susceptible to adversarial attacks \cite{szegedy2014intriguing,goodfellow2014explaining}. A line of empirical defenses~\cite{buckman2018thermometer,song2018pixeldefend} have been proposed to defend adversarial attacks, but are often broken by the newly devised stronger attacks~ \cite{RN7}. Existing certified defenses ~ \cite{RN9,RN22,RN2} provide the theoretical guarantees for their robustness. In particular, \emph{Randomized smoothing} \cite{RN2} is one of the few certified defenses that can scale to ImageNet-scale classification task, showing its great potential for wide application. Moreover, randomized smoothing has shown high robustness against various types of adversarial attacks, including norm-constrained perturbations (e.g. $\ell_0, \ell_2, \ell_\infty$ norms) and image transformations (e.g. rotations and image shift). 

\begin{figure}[t!]
    \begin{subfigure}{.5\textwidth}
    \centering
    \includegraphics[width=0.8\linewidth]{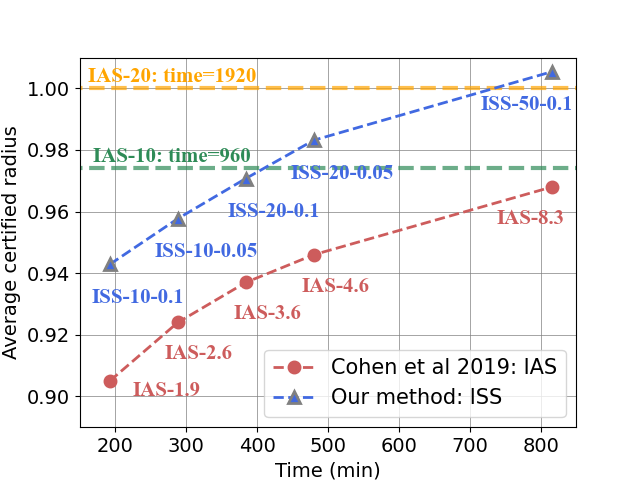}
  \end{subfigure}%
    \vspace{-5pt}
  \caption{ISS achieves a better ACR-time trade-off than IAS.
  $\rm{ISS}-c_1-c_2$ denotes that ISS accelerates $\rm{IAS}-c_1$ at the controllable certified radius decline ($\leq c_2$). $\rm{IAS}-c_1$ denotes that IAS accelerates $\rm{IAS}-10, \rm{IAS}-20$ by reducing the sample size to $c_1 \times 10,000$. ISS always surpasses IAS on ACR in the same time cost. The results are evaluated on the ImageNet ($\sigma=1.0$) model trained by \cite{DBLP:conf/nips/JeongS20}.}
  \label{fig:intro}
  \vspace{-10pt}
\end{figure}

Despite these advances, randomized smoothing suffers the costly robustness certification. Specifically, computing a certified radius close to the exact value needs a relatively tight lower bound of the top-1 label probability, which requires running forward passes on a large number of samples \cite{RN1, RN2, DBLP:conf/iclr/ZhaiDHZGRH020, DBLP:conf/nips/JeongS20,RN51}. Such expensive overheads make them less applicable to the real-world scenarios. Some works \cite{RN51, DBLP:conf/aaai/FengWCZN20} proposed to leverage the runner-up label probability in the certification, but their performances may suffer from the inevitable loss in the simultaneous confidence intervals. Traditionally, the robustness certification is accelerated by reducing the sample size used for estimating the lower bound ~\cite{RN2,RN51}, but the vanilla sample size reduction will lead to a poor ACR-runtime trade-off. It is critical to develop a cost-effective certification method.



In this paper, we propose Input-Specific Sampling (ISS) to speed up the certification for randomized smoothing, without hurting too much on the certification performance. The idea behind ISS is to \emph{minimize the sample size for the given input at the bounded cost of the certified radius decline,} instead of directly applying the same sample size to all inputs. The idea is realized by precomputing a mapping from the input characteristics to the sample size. Consequently, ISS can accelerate the certification at a controllable cost. Empirical results validate that ISS consistently outperforms IAS \cite{RN2} on ACR. As shown in Fig. \ref{fig:intro}, $\rm{ISS}-10-0.05$ (ACR=0.958) accelerates the standard certification $\rm{IAS}-10$, shortening the certification time $962 \rightarrow 250$ mins at the controllable decline ($\leq 0.05$). Furthermore, ISS is compatible with all the randomized smoothing works that need confidence interval, since ISS has no additional constraint on the base classifier or the smoothing scheme.

Our contributions can be summarized as follows;

\begin{enumerate}
    \item We propose Input-Specific Sampling (ISS) to adaptively reduce the sample size for each input. The proposed input-specific sampling, for the first time to our best knowledge, can significantly reduce the cost for accelerating the robustness certification of randomized smoothing.
    
    
    \item ISS can universally control the difference between the certified radii before and after the acceleration. In particular, the sample size computed by ISS is theoretically tight for bounding the radius decline.
    
    
    
    
    
    \item The results on CIFAR-10 and ImageNet demonstrate that: 1) ISS significantly accelerates the certification at a controllable decline in the certified radii. 2) ISS consistently achieves a higher average certified radius when compared to the mainstream acceleration IAS.
\end{enumerate}


\section{Related Works}
\label{sec:related}
\paragraph{Certified defenses.}
Neural networks are vulnerable to adversarial attacks \cite{RN7,RN45, Kurakin2017AdversarialML,RN45,jia2018attriguard}.
Compared to empirical defenses \cite{goodfellow2014explaining,svoboda2018peernets,buckman2018thermometer,ma2018characterizing,guo2018countering,dhillon2018stochastic,RN33,song2018pixeldefend},
certified defenses can provide provable robustness guarantees for their predictions. Recently, a line of certified defenses have been proposed, including dual network ~\cite{RN9}, convex polytope ~\cite{RN26}, CROWN-IBP ~\cite{RN35}, Lipschitz bounding ~\cite{RN11}. However, those certified defenses suffer from either the low scalability or the hard constraints on the neural network architecture. 

\paragraph{Randomized smoothing.}
In the seminal work \cite{RN2}, the authors for the first time propose randomized smoothing to defend the $\ell_2$-norm perturbations, which significantly outperforms other certified defenses. Recently, series of works further extend randomized smoothing to defend various attacks, including $\ell_0, \ell_1, \ell_2, \ell_\infty$-norm perturbations and geometric transformations. For instance, \cite{levine2020robustness} introduce the random ablation against $\ell_0$-norm adversarial attacks. \cite{yang2020randomized} propose Wulff Crystal uniform distribution against $\ell_1$-norm perturbations. \cite{Awasthi2020low} introduce $\infty \rightarrow 2$ matrix operator for Gaussian smoothing to defend $\ell_\infty$-norm perturbations. \cite{fischer2020certified,Li2020ProvableRL} exploit randomized smoothing to defend adversarial translations. Remarkably, almost all the randomized smoothing works \cite{RN1, RN2, DBLP:conf/iclr/ZhaiDHZGRH020, DBLP:conf/nips/JeongS20, yang2020randomized, RN51} have achieved superior certified robustness to other certified defenses in their respective fields. 

\paragraph{Robustness certification in randomized smoothing.}
Despite its sound performance, the certification of randomized smoothing is seriously costly. Unfortunately, accelerating the certification is a fairly under-explored field. The mainstream acceleration method \cite{RN51,DBLP:conf/aaai/FengWCZN20}, which we call IAS, is to apply a smaller sample size for certifying the radius. However, IAS accelerates the certification at a seriously sacrifice ACR and the certified radii of specific inputs. Therefore, it calls for approaches to achieve a better time-cost trade-off, which is the main purpose of this paper.

\section{Preliminaries}
\label{sec:pre}
\paragraph{Randomized smoothing}
The basic idea of randomized smoothing \cite{RN2} is to generate a smoothed version of the base classifier $f$. Given an arbitrary base classifier $f(x): \R^d \rightarrow \y$ where $\y=\{1,\dots,k \}$ is the output space, the smoothed classifier $g(\cdot)$ is defined as:
\begin{equation}
\begin{split}
    &g(x) \defeq \argmax_{c \in \y} \r{} [f(x')=c],\; x' \sim \mathcal{N}(x,\sigma^2 I^d)
\end{split}
\end{equation}
$g(x)$ returns the most likely predicted label of $f(\cdot)$ when input the data with Gaussian augmentation $x' \sim \mathcal{N}(x,\sigma^2 I^d)$. The tight lower bound of $\ell_2$-norm certified radius \cite{RN2} for the prediction $c_A=g(x)$ is:
\begin{equation}
\begin{split}
\label{eq:1}
   & \sigma \Phi^{-1} (p_A)\\
   & \text{ where } p_A \defeq \r{} [f(x')=c_A],\; x' \sim \mathcal{N}(x,\sigma^2 I^d)
\end{split}
\end{equation}
where $\Phi^{-1}$ is the inverse of the standard Gaussian CDF. We emphasize that computing the deterministic value of $g(x)$ is impossible because $g(\cdot)$ is built over the random distribution $\mathcal{N}(x,\sigma^2 I^d)$. Therefore, we use Clopper-Pearson method \cite{clopper1934use} to guarantee $\r{} [f(x')=c_A] > \r{} [f(x')=c], \forall c \neq c_A$ with the confidence level $1-\alpha$, and then we have $g(x)=c_A$ with the confidence level $1-\alpha$.

\paragraph{Robustness certification} In practice, the main challenge in computing the radius $\sigma \Phi^{-1} (p_A)$ is that $p_A$ is inaccessible because iterating all possible $f(x'): x'\in \R^d$ is impossible. Therefore, we estimate $\underline{p_A}$, the standard one-sided Clopper-Pearson confidence lower bound of $p_A$ instead of $p_A$ and certify a lower bound $\sigma \Phi^{-1} (\underline{p_A})$. Estimating a tight $\underline{p_A}$ needs a large size of samples for $f(x'): x' \sim \mathcal{N}(x,\sigma^2 I^d)$. Generally, the estimated $\underline{p_A}$ increases with the sample size\footnote{The seminal work \cite{RN2} derives the certified radius: $\frac{\sigma}{2}[\Phi^{-1}(p_A)-\Phi^{-1}(p_B)]$ where $p_B$ is the runner-up label probability. Currently, most works \cite{RN2, DBLP:conf/iclr/ZhaiDHZGRH020, DBLP:conf/nips/JeongS20, RN51} compute the certified radius by Eq. (\ref{eq:1}), which substitutes $p_B$ with $1-p_A$, to avoid doing interval estimation twice.}.


\paragraph{Standard certification and vanilla acceleration: IAS} 
The standard certification algorithm \cite{RN2} can be summarized in two steps:
\begin{enumerate}
    \item \textbf{Sampling:} Given the input $x$, sample $k$ (e.g. $k=100,000$) iid samples $\{x'_i: i=1,\dots,k\} \sim \mathcal{N}(x,\sigma^2 I^d)$ and run $k$ times forward passes $\{f(x'_i):i=1,\dots,k\}$.
    \item \textbf{Interval estimation:} Count $k_A = \sum_{i=1}^k \i\{f(x'_i)=c_A\}$ ($\i$ denotes the indicator function) where $c_A$ is the label with top-1 label counts. Compute the confidence lower bound $\underline{p_A}$ with the confidence level $1-\alpha$. Return the certified radius $\sigma \Phi^{-1}(\underline{p_A})$.
\end{enumerate}
The high computation is mainly due to the $k$ times forward passes in \textbf{Sampling}. The certification is accelerated by the vanilla sample size reduction, which we call input-agnostic sample size reduction (IAS). This acceleration is at the cost of unpredictable radius declines, which yields a poor ACR-runtime trade-off since it reduces the sample size equally for each input, without considering the input characteristics. 

\section{Methodology}
\label{sec:method}
\begin{figure*}[tb]
    \centering
    \includegraphics[width=0.9\linewidth]{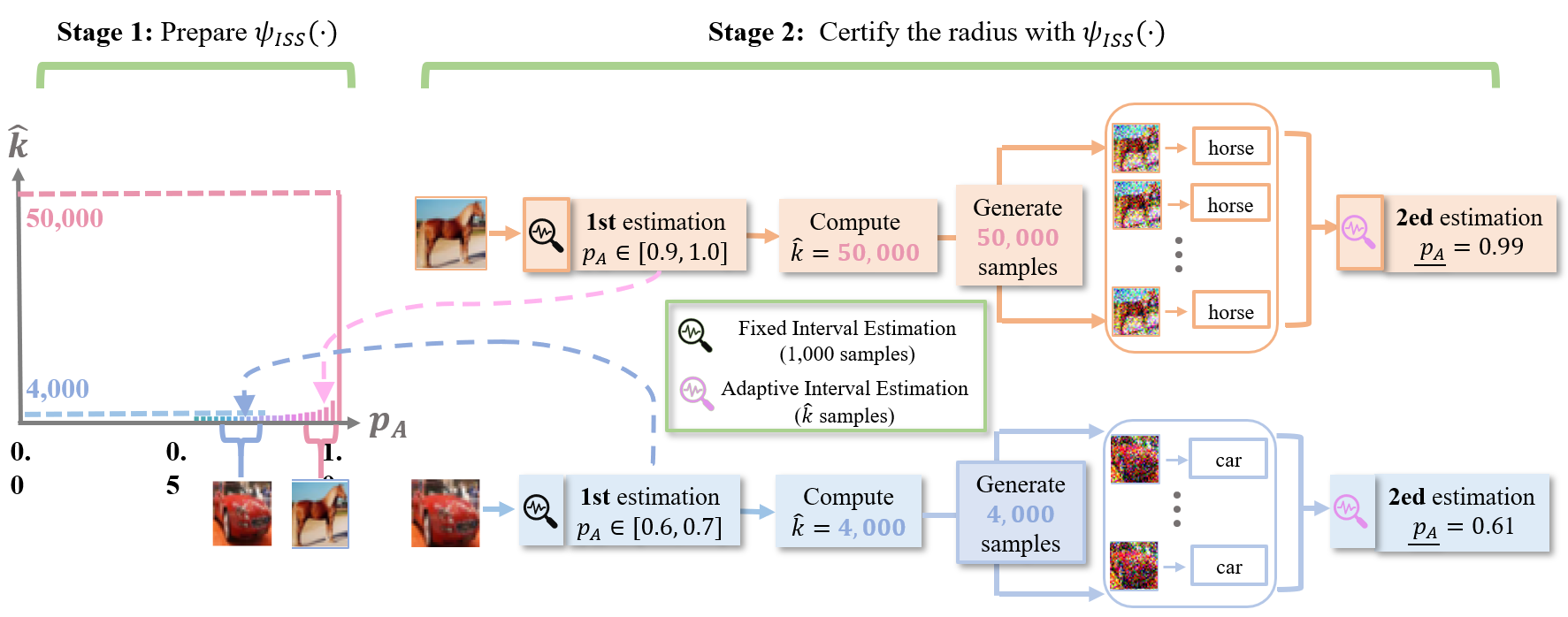}
    \vspace{-10pt}
    \caption{Overview of the robustness certification with ISS. In Stage 1, we compute $\psi_{ISS}(\cdot)$, a mapping from $p_A$ to $\hat{k}$. In Stage 2, given the image $x$, we first loosely estimate the confidence interval for $p_A$ and determine the sample size for $x$ with $\psi_{ISS}(\cdot)$. }
    \label{fig:overview}
     \vspace{-5pt}
\end{figure*}

We first introduce the notions of Absolute Decline and Relative Decline. Then we propose Input-Specific Sampling (ISS), which aims to use the minimum sample size with the constraint that the radius decline is less than the given bound.

\subsection{Overview and main idea} 
The key idea of ISS is to appropriately reduce the sample size for each input, instead of applying the same sample size to the certifications for all inputs. Since the sample size reduction will inevitably cause the decline in the certified radius, thus we aim to quantify the radius decline and bound the decline to be less than the pre-specified value. First we define the radius decline as follows:
\begin{definition}[\textbf{Absolute Decline $\mathrm{AD}(k;\overline{k},p)$}]
Given the input $x$ and the pre-specified desired sample size $\overline{k}$ (e.g. $\overline{k}=100,000$), suppose we know $p_A$ of $x$, Absolute Decline $\mathrm{AD}(k;\overline{k},p)$ is the gap between the radius certified at the sample size $\overline{k}$ and the radius certified at ${k}: k \leq \overline{k}$: 
\begin{equation}
\begin{split}
    &\mathrm{AD}(k;\overline{k},p_A) \defeq \underbrace{\sigma \Phi^{-1}(\underline{p_1})}_{\text{Desired radius}} - \underbrace{\sigma \Phi^{-1}(\underline{p_2})}_{\text{Estimated radius}} \\
    \text{where }& \underline{p_1}=\mathbf{B}(\alpha;\;p_A \overline{k},\;\overline{k}-p_A \overline{k}+1 ),\\ 
    & \underline{p_2}=\mathbf{B}(\alpha;\; p_A k,\;k- p_A k+1 )
\end{split}
\end{equation}
where $\mathbf{B}(\alpha; k_A, k-k_A+1)$ denotes the one-sided Clopper-Pearson lower bound \cite{clopper1934use} with the confidence level $1-\alpha$, which is equal to the $\alpha$th quantile from a Beta distribution with shape parameters $k_A, k-k_A+1$.
\end{definition}
\begin{definition}[\textbf{Relative Decline} $\mathrm{RD}(k;\overline{k},p_A)$]
Similar to absolute decline, Relative Decline $\mathrm{RD}(k;\overline{k},p_A)$ is
\begin{equation}
\begin{split}
        &\mathrm{RD}(k;\overline{k},p_A) \defeq \frac{\sigma \Phi^{-1}(\underline{p_1})-\sigma \Phi^{-1}(\underline{p_2})}{ \sigma \Phi^{-1}(\underline{p_1})}\\
       \text{where }& \underline{p_1}=\mathbf{B}(\alpha;\;p_A \overline{k},\;\overline{k}-p_A \overline{k}+1 ),\\ 
    & \underline{p_2}=\mathbf{B}(\alpha;\; p_A k,\;k- p_A k+1 )
\end{split}
\end{equation}
\end{definition}
\begin{remark}
The absolute (or relative) decline is the expected gap between the radius certified at the sample size $\overline{k}$ and $k$ when fixing $k_A/k \equiv p_A$ where $k_A \defeq \sum_{i=1}^k \mathbb{I}\{f(x'_i)=c_A\}$. It connects the expected radius decline to the sample size when given $p_A$. In particular, when $\overline{k}=\infty$, the absolute (or relative) decline measures the gap between the optimal certified radius that randomized smoothing can provide and the radius certified at the sample size $k$.
\end{remark}
\paragraph{Formulate our key idea} Given the input $x$ and the pre-specified upper bound of the decline $\rm{U}_{AD} \in \mathbb{R}^+$ (or $\rm{U}_{RD} \in \mathbb{R}^+$), our idea for $\rm{AD}$ (or $\rm{RD}$) is formulated as follows: 
\begin{enumerate}
    \item find $\min k$ with the constraint $\mathrm{AD}(k;\overline{k},p) \leq \rm{U}_{AD}$.
    \item find $\min k$ with the constraint $\mathrm{RD}(k;\overline{k},p) \leq \rm{U}_{RD}$.
\end{enumerate}
In practice, solving the above two problems is non-trival because $p_A$ of $x$ is inaccessible. Simply treating the estimated $k_A/k$ as $p_A$ is obviously unreasonable. We propose ISS, a practical solution to the above two problems.

\begin{algorithm}[t!]
\caption{Compute ISS mapping $\psi_{ISS}(\cdot)$}
\label{alg:ssp}
\textbf{Input}: The maximum decline $\rm{U}$, the decline type, the desired sample size $\overline{k}$, the noise level $\sigma$, the confidence level $1-\alpha$, the length of the subinterval $\delta$ \\
\textbf{Output}: the ISS mapping $\psi_{ISS}(\cdot)$
\begin{algorithmic}[1]
\FOR{$N = 0, 1, 2, \dots, \frac{1}{\delta}$}
\STATE $p \leftarrow N \cdot \delta$;
\STATE Compute
$\overline{r} = \sigma \cdot \Phi^{-1}\left(\rm{B}(\alpha;\; p \overline{k},\; \overline{k})\right)$;
\STATE Compute the minimum required certified radius:\\
$\text{If the decline type is $\rm{AD}$:}\; \tilde{r} \leftarrow \overline{r}-\rm{U}_{AD} \quad$ or\\
$\text{If the decline type is $\rm{RD}$:}\; \tilde{r} \leftarrow (1-\rm{U}_{RD})\overline{r}$;
\IF{$\tilde{r} \leq 0$}
\STATE $\psi_{ISS}(p) \leftarrow 0$;
\ELSE
\STATE $\psi_{ISS}(p) \leftarrow \argmin\limits_k \sigma \cdot \Phi^{-1} \left(\rm{B}(\alpha;\; p k,\; k)\right) \geq \tilde{r}$;
\ENDIF
\ENDFOR
\STATE \textbf{Return} $\psi_{ISS}(p): p=\delta, 2 \delta, \dots, 1$;
\end{algorithmic}
\end{algorithm}

\begin{algorithm}[t!]
\caption{Certification with input-specific sampling (ISS)}
\label{alg:certify}
\textbf{Input}: The input $x$, the base classifier $f$, the maximum sample size $\overline{k}$, the sample size $k_0: k_0 \leq \overline{k}$, the confidence level $\alpha$, the ISS mapping $\psi_{ISS}(\cdot)$\\
\textbf{Output}: Prediction $pred$, radius $r$
\begin{algorithmic}[1]
\STATE Sample $k_0$ noisy samples $x'_1,\dots,x'_k \sim \mathcal{N}(x,\sigma^2 I^d)$;
\STATE Compute the prediction:\\
$pred \leftarrow \argmax_{y \in \y} \sum_{i=1}^{k_0} \mathbb{I}\{f(x'_i)=y\}$;
\STATE Count $k^0_A \leftarrow \max_{y \in \y} \sum_{i=1}^{k_0} \mathbb{I}\{f(x'_i)=y\}$;
\STATE Compute the two-sided confidence interval:\\
$p_{\rm{low}} \leftarrow \mathbf{B}(\alpha/2;\;k^0_A,\;k_0-k^0_A+1)$\\
$p_{\rm{up}} \leftarrow \mathbf{B}(1-\alpha/2;\;k^0_A+1,\;k_0-k^0_A)$;
\STATE Compute $\hat{k} \leftarrow \max(\psi_{ISS}(p_{\rm{low}}), \psi_{ISS}(p_{\rm{up}}))$; \\
\STATE Sample $\max(\hat{k}-k_0,0)$ noisy samples:\\ $x'_{k_0+1},\dots,x'_{\hat{k}} \sim \mathcal{N}(x,\sigma^2 I^d)$;
\STATE Count $k_A \leftarrow \max_{y \in \y} \sum_{i=1}^{\hat{k}} \mathbb{I}\{f(x'_i)=y\}$;
\STATE Compute the one-sided confidence lower bound:\\
$\underline{p_A} \leftarrow \mathbf{B}(\alpha;\;k_A,\;\hat{k}-k_A+1)$;
\IF{$\underline{p_A}<\frac{1}{2}$}
\STATE $pred \leftarrow \rm{ABSTAIN}, r \leftarrow 0$;
\ELSE
\STATE Compute the radius $r \leftarrow \sigma \Phi^{-1}(\underline{p_A})$;
\ENDIF
\STATE \textbf{Return} $pred$ and $r$;
\end{algorithmic}
\end{algorithm}

\subsection{Certification with input-specific sampling}
Fig. \ref{fig:overview} shows an overview. Given the input $x$, we first estimate a relatively loose two-sided Clopper-Pearson confidence interval $p_A \in [p_{\rm{low}}, p_{\rm{up}}]$ by $k_0$ samples where $k_0 < \overline{k}$ is a relatively small sample size. Given $\overline{k}, \rm{U}_{AD}$ (or $\rm{U}_{AD}$), ISS assigns the sample size $\hat{k}$ for certifying $g(x)$ where $\hat{k}$ is:
\begin{align}
\label{opt:1}
        \hat{k}=  & \max \left(\psi(p_{\rm{low}}), \psi(p_{\rm{up}}) \right)\\
\text{For Absolute Decline}: & \; \psi(p) \defeq \argmin_k \; \rm{AD}(k;\overline{k},p) \leq \rm{U}_{AD} \nonumber\\
\text{For Relative Decline}: & \; \psi(p) \defeq \argmin_k \; \rm{RD}(k;\overline{k},p) \leq \rm{U}_{RD} \nonumber
\vspace{-10pt}
\end{align}

Formally, we present the following two propositions to theoretically prove that $\hat{k}$ ($\rm{AD}$) computed from Eq. (\ref{opt:1}) is optimal. Prop. \ref{pro:1} guarantees that the sample size $\hat{k}$ computed from Eq. (\ref{opt:1}) must satisfy the constraint $\mathrm{AD}(\hat{k};\overline{k},p_A) \leq \rm{U}_{AD}$. Prop. \ref{pro:2} guarantees that $\hat{k}$ is the minimize sample size that can guarantee $\mathrm{AD}(\hat{k};\overline{k},p_A) \leq \rm{U}_{AD}$.

\begin{proposition}
\label{pro:1}
[\textbf{Bounded absolute radius decline}] Suppose $p_A \in [p_{\rm{low}}, p_{\rm{up}}]$ with $1-\alpha$ confidence level, then we guarantee that there is at least $1-\alpha$ probability that  $\hat{k}$ computed from Eq. (\ref{opt:1}) satisfies $\rm{AD}(k; \hat{k}, p_A) \leq \rm{U}_{AD}$.
\end{proposition}

\begin{proposition}
\label{pro:2}
[\textbf{Tightness for $\hat{k}$}] Suppose $p_A \in [p_{\rm{low}}, p_{\rm{up}}]$ and $\hat{k}$ is computed from Eq. (\ref{opt:1}), then for an arbitrary sample size $k: k < \hat{k}$, there exists $p_A \in [p_{\rm{low}}, p_{\rm{up}}]$ that breaks the constraint $\rm{AD}(k; \hat{k}, p_A) \leq \rm{U}_{AD}$.
\end{proposition}



\subsection{Implementation}
 In the practical algorithm of ISS, we substitute $\psi(\cdot)$ in Eq. (\ref{opt:1}) with a piecewise constant function approximation $\psi_{ISS}(\cdot)$. The advantage of $\psi_{ISS}(\cdot)$ over $\psi(\cdot)$ is that we can compute $\psi_{ISS}(p): p\in [0,1]$ previously before the certification to save the cost in computing $\psi(p_{\rm{low}}), \psi(p_{\rm{up}})$ in Eq. (\ref{opt:1}) when certifying the radius for the testing data. Constructing $\psi_{ISS}(\cdot)$ is feasible because that the value of $\psi(p)$ only depends on $p$ when fixing $\overline{k}$, regardless of the testing set or the base classifier architecture. Specifically, $\psi_{ISS}(p)$ is
\begin{small}
\begin{align}
\label{eq:piecewise}
    \psi_{ISS}(p)=
&\begin{cases}
    \psi(p) & p / \delta \in \mathbb{N} \\
    \max(\psi(N_1 \delta), \psi(N_2 \delta) ) & p / \delta \in (N_1 , N_2)
    \end{cases}
\end{align}
\end{small}
where $N_1, N_2 \in \mathbb{N}$. Obviously, $\forall p \in [0,1], \psi_{ISS}(p) \geq \psi(p)$, thus Prop. \ref{pro:1} still holds for the substitution $\psi_{ISS}(\cdot)$. Prop. \ref{pro:2} holds for $\psi_{ISS}(\cdot)$ when $p_{\rm{low}} / \delta \in \mathbb{N}, p_{\rm{up}} / \delta \in \mathbb{N}$.
 
The practical algorithm can summarized into two stages:
 \vspace{5pt}
 
\noindent\textbf{Stage 1: prepare $\psi_{ISS}(\cdot)$.} Given $\overline{k}$ and the decline upper bound $\rm{U}_{AD}$ (or $\rm{U}_{RD}$), compute $\psi_{ISS}(p)$ by Eq. (\ref{opt:1}) and Eq. (\ref{eq:piecewise}). The detailed algorithm is shown in Alg. \ref{alg:ssp}.
 \vspace{5pt}
\noindent\textbf{Stage 2: certify the radius with $\psi_{ISS}(\cdot)$.} Given $x$, we first estimate a loose confidence interval  $p_A \in [p_{\rm{low}}, p_{\rm{up}}]$ by $k_0$ samples. With $[p_{\rm{low}}, p_{\rm{up}}]$ and $\psi_{ISS}(\cdot)$, we compute the input-specific sample size $\hat{k}$. Then we estimate the certified radius by sampling $\hat{k}$ noisy samples. The algorithm is shown in Alg. \ref{alg:certify}.



\begin{figure}[t!]
\centering
  \begin{subfigure}[b]{.25\textwidth}
  \centering
    \includegraphics[width=0.75\linewidth]{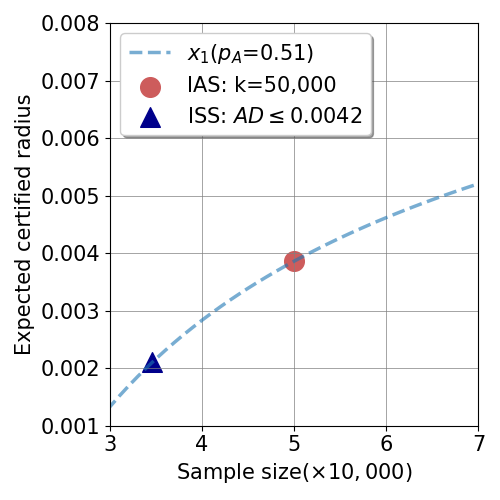}
    \vspace{-5pt}
    \caption{$\hat{R}$-$k$ curve of $x_1$ ($p_A=0.51$).}
    \label{fig:x1}
  \end{subfigure}%
    \begin{subfigure}[b]{.25\textwidth}
    \centering
    \includegraphics[width=0.75\linewidth]{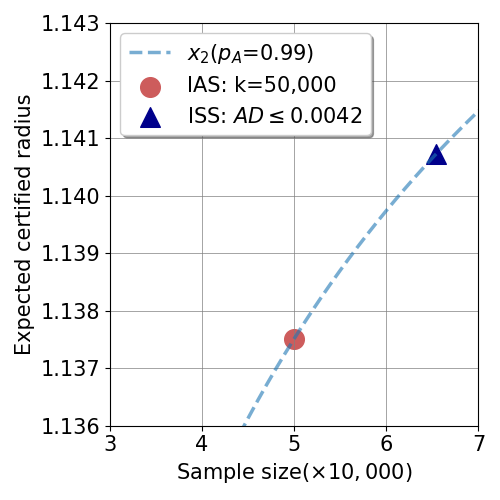}
    \vspace{-5pt}
    \caption{$\hat{R}$-$k$ curve of $x_2$ ($p_A=0.99$).}
    \label{fig:x2}
  \end{subfigure}%
   \vspace{-5pt}
   \caption{ISS certifies higher ACR on $x_1, x_2$ than IAS.}
   \label{fig:rdad-pa}
   \vspace{-10pt}
\end{figure}

\begin{figure}[t!]
\centering
  \begin{subfigure}[b]{.25\textwidth}
  \centering
    \includegraphics[width=0.75\linewidth]{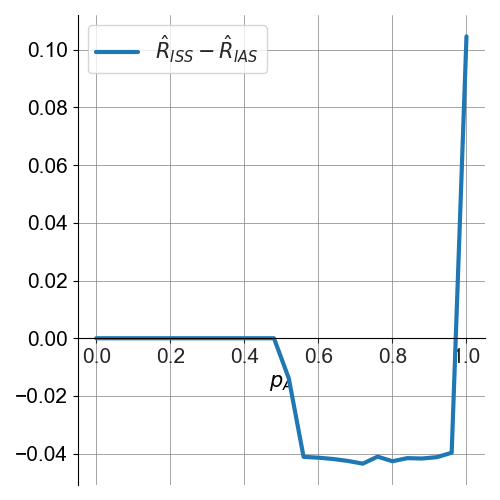}
    \vspace{-5pt}
    \caption{$\hat{R}_{ISS}-\hat{R}_{IAS}$ w.r.t. $p_A$}
    \label{fig:compare}
  \end{subfigure}%
    \begin{subfigure}[b]{.25\textwidth}
    \centering
    \includegraphics[width=0.75\linewidth]{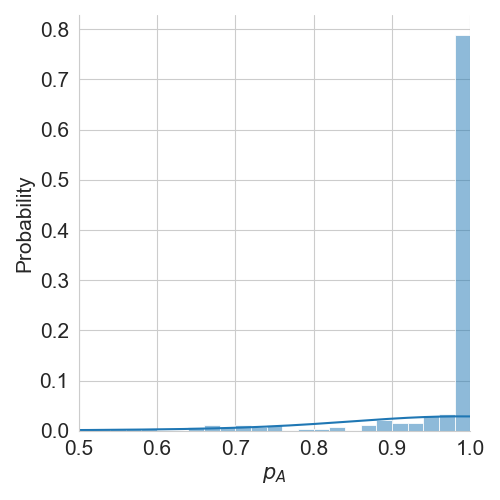}
    \vspace{-5pt}
    \caption{$p_A$ distribution (ImageNet).}
    \label{fig:pa_dist}
  \end{subfigure}%
  \vspace{-5pt}
   \caption{ISS fits the practical smoothed classifiers.}
   \vspace{-10pt}
\end{figure}

\paragraph{Compare ISS(AD) to IAS} We compare ISS to IAS in Fig. \ref{fig:x1}, Fig. \ref{fig:x2} where $\hat{R}(k, p_A; \sigma) \defeq \sigma \Phi^{-1}(\mathbf{B}(\alpha; p_A k,k- p_A k+1))$. As presented, IAS assigns $50,000$ for both $x_1, x_2$, while ISS assigns $35,000$ and $65,000$ for $x_1, x_2$ respectively. The sample size of ISS are computed by solving $\rm{AD}(k;100,000,p_A) \leq 0.0042$. For each certified radius, the decline in $x_2$ certified radius is up to $0.0075$ due to the sample size reduction $100,000 \rightarrow 50,000$ , which is $1.78 \times$ $\rm{U}_{AD}$ of ISS. For the average certified radius, ISS trades $0.002$ radius of $x_1$ for $0.003$ radius of $x_2$, thus ACR of ISS improves IAS $0.0005$ under the same average sample size. The improvement is because ISS tends to assign larger sample sizes to the high-$p_A$ inputs, which meets the property of $\hat{R}(k, p_A; \sigma)$. Namely, $\hat{R}(k+\Delta k, p_A; \sigma)-\hat{R}(k, p_A; \sigma)$ increases with $p_A$, meaning that assigning larger sample sizes to the high-$p_A$ inputs is more efficient than input-agnostic sampling. 
\vspace{-5pt}
\paragraph{ISS fits the well-trained smoothed classifiers} Fig. \ref{fig:compare} reports $\hat{R}_{ISS}-\hat{R}_{IAS}$ where $\hat{R}_{ISS}$ denotes the radius certified by ISS of $\overline{k}=100,000, \rm{U}_{AD}= 0.05$ and $\hat{R}_{IAS}$ denotes the radius certified by IAS at $k=30,000$\footnote{Here we choose to compare ISS to IAS ($k=30,000$) is because that the average sample size of ISS ($\overline{k}=100,000, \rm{U}_{AD}= 0.05$) on the ImageNet model trained by Consistency \cite{DBLP:conf/nips/JeongS20} ($\sigma=0.5$) is roughly $30,000$.}. We observe that ISS certifies higher certified radius when $p_A > 0.94$. Fig. \ref{fig:pa_dist} reports the $p_A$ distribution of the test set \footnote{We sample $k=1000,000$ Monte Carlo samples and approximately regard $k_A /k$ as the exact value of $p_A$.} on the real ImageNet base classifier ($\sigma=0.5$) trained by Consistency \cite{DBLP:conf/nips/JeongS20}. We found that the probability mass of $p_A$ distribution is concentrated around $p_A=1.0$, which is the interval where $\hat{R}_{ISS}-\hat{R}_{IAS} >0$. Furthermore, ISS is expected to outperform IAS on the smoothed classifiers trained by other algorithms, since their $p_A$ distributions have the similar property (see appendix).




\section{Experiments}
\label{sec:exper}
We evaluate our proposed method ISS on two benchmark datasets: CIFAR-10 \cite{dataset/cifar} and ImageNet \cite{ILSVRC15}. All the experiments are conducted on CPU (16 Intel(R) Xeon(R) Gold 5222 CPU @ 3.80GHz) and GPU (one NVIDIA RTX 2080 Ti). We observe that the certification runtime is roughly proportional to the average sample size when fixing the model architecture, as shown in Table \ref{tab:imagenet_ard}. The hyperparameters are listed in Table \ref{tab:config}. For clarity, $\rm{ISS}-c_1-c_2$ denotes ISS at $\overline{k}=c_1 \cdot 10,000, \rm{U}_{AD}=c_2$, and $\rm{IAS}-c_1$ denotes IAS at $k=c_1 \cdot 10,000$. The overhead of computing $\psi_{ISS}$ is reported in Table \ref{tab:time}.

\subsection{Evaluation metrics}
Our evaluation metrics include average sample size, runtime, MAD, ACR and certified accuracy, where MAD denotes the maximum absolute decline between the radius certified before and after the acceleration among all the testing data\footnote{Note the speedup of ISS deterministically depends on the $p_A$ distribution of the testing set. Since the smoothed classifiers trained by different training algorithms, including SmoothAdv \cite{RN1}, MACER \cite{DBLP:conf/iclr/ZhaiDHZGRH020} and Consistency \cite{DBLP:conf/nips/JeongS20}, report the similar $p_A$ distributions, ISS will perform similarly on the models trained by other algorithms.}. ACR and certified accuracy $\rm{CA}(r)$ at the radius $r$ are computed as follows:
\begin{small}
\begin{align}
    &\rm{ACR} \defeq \frac{1}{|\mathcal{D}_{test}|} \sum_{(x,y) \in \mathcal{D}_{test}} R(x;g) \cdot \mathbb{I}(g(x)=y) \\
    &\rm{CA}(r) \defeq  \frac{1}{|\mathcal{D}_{test}|} \sum_{(x,y) \in \mathcal{D}_{test}} \mathbb{I}(R(x;g)>r) \cdot \mathbb{I}(g(x)=y)
\end{align}
\end{small}
where $\rm{R}(x;g)$ denotes the estimated certified radius of $g(x)$. 



\begin{figure*}[t!]
\centering
   \begin{subfigure}[b]{.24\textwidth}
  \centering
    \includegraphics[width=1.0\linewidth]{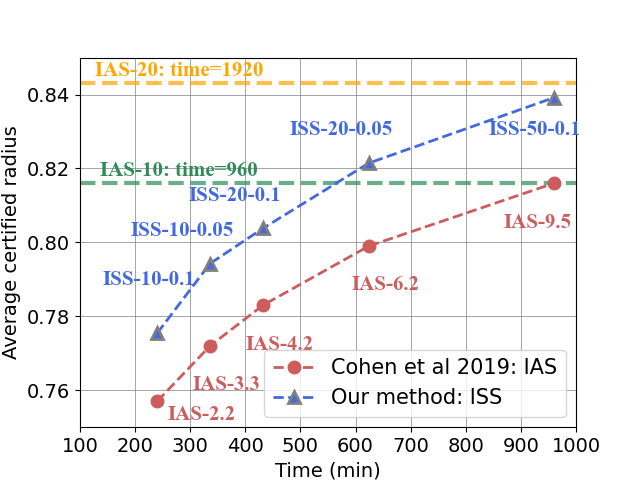}
    \caption{ImageNet ($\sigma=0.5$).}
    \label{fig:img_5}
  \end{subfigure}%
    \begin{subfigure}[b]{.24\textwidth}
    \centering
    \includegraphics[width=1.0\linewidth]{fig/img_1.0.png}
    \caption{ImageNet ($\sigma=1.0$).}
    \label{fig:img_10}
  \end{subfigure}
  \vspace{-10pt}
  \begin{subfigure}[b]{.24\textwidth}
  \centering
    \includegraphics[width=1.0\linewidth]{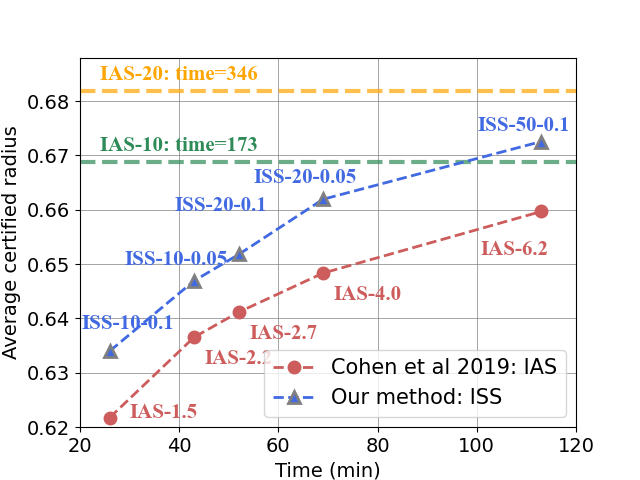}
    \caption{CIFAR-10 ($\sigma=0.5$).}
    \label{fig:cifar_5}
  \end{subfigure}%
    \begin{subfigure}[b]{.24\textwidth}
    \centering
    \includegraphics[width=1.0\linewidth]{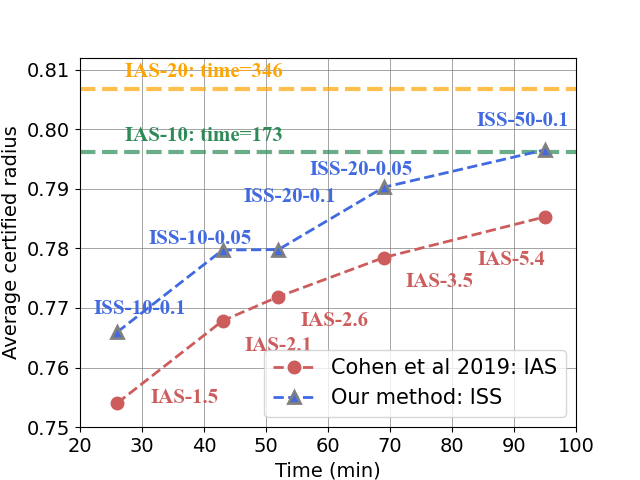}
    \caption{CIFAR-10 ($\sigma=1.0$).}
    \label{fig:cifar_10}
  \end{subfigure}%
      \caption{Overall analysis on ImageNet and CIFAR-10.}
      \vspace{-5pt}
\end{figure*}
\begin{figure*}[tb!]
\centering
  \begin{subfigure}[t!]{.24\textwidth}
  \centering
    \includegraphics[width=1.0\linewidth]{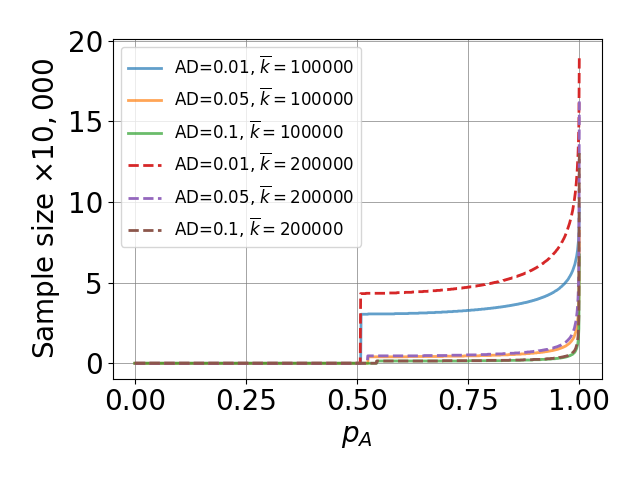}
  \end{subfigure}%
    \begin{subfigure}[t!]{.24\textwidth}
    \centering
    \includegraphics[width=1.0\linewidth]{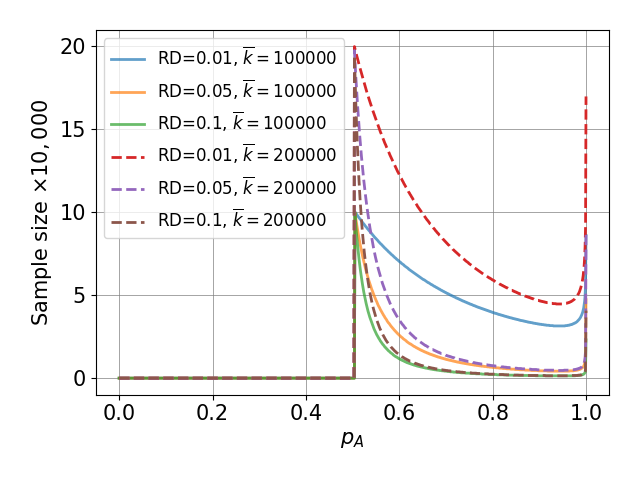}
  \end{subfigure}%
  \begin{subfigure}[t!]{.24\textwidth}
    \centering
    \includegraphics[width=1.0\linewidth]{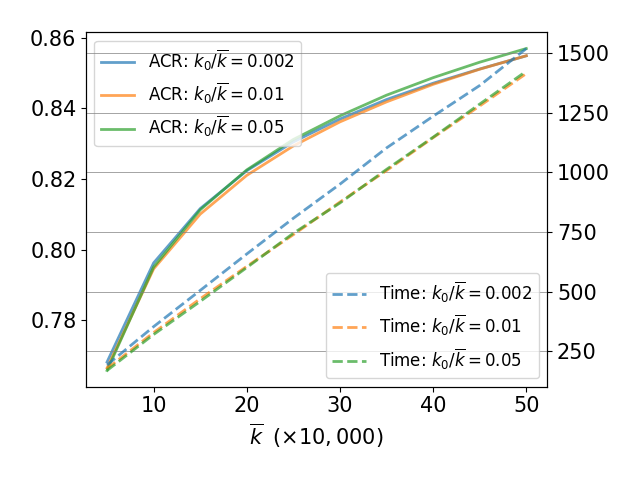}
  \end{subfigure}%
  \begin{subfigure}[t!]{.24\textwidth}
    \centering
    \includegraphics[width=1.0\linewidth]{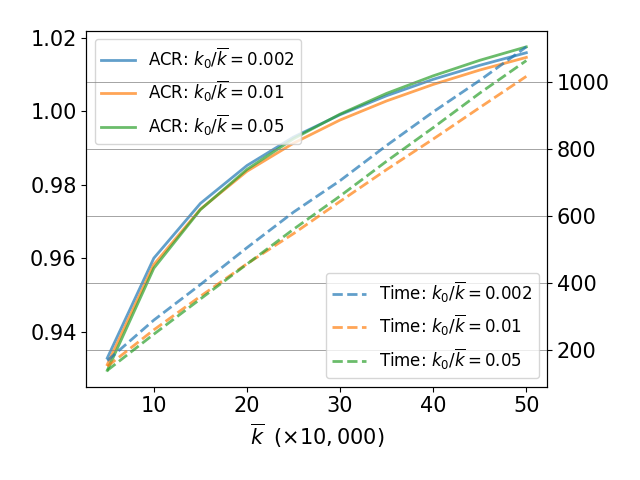}
  \end{subfigure}
   \vspace{-15pt}
   \caption{Ablation studies. \textbf{Upper:} $k$-$p_A$ curves w.r.t. $\rm{AD}, \overline{k}$ (\textbf{Upper Left}) and $\rm{RD}, \overline{k}$ (\textbf{Upper Right}). \textbf{Lower:} ACR-$\overline{k}$ curves and Time-$\overline{k}$ curves w.r.t. $k_0/\overline{k}$ on ImageNet $\sigma=0.5$  (\textbf{Lower Left}) and ImageNet $\sigma=1.0$ (\textbf{Lower Right}).} 
    \label{fig:ablation}
\end{figure*}
\begin{table*}[t!]
\centering
\caption{ImageNet: compare ISS to IAS on average sample size (Avg), certification runtime, maximum absolute decline (MAD), verage certified radii (ACR) and certified accuracies ($\%$) on the models trained by Consistency \cite{DBLP:conf/nips/JeongS20}. Results are evaluated on $500$ trials by $id=0, 100, \dots, 49800, 49900$. The bold denotes better performance under the similar setting.} 
\vspace{-10pt}
\label{tab:imagenet_ard}
    \begin{adjustbox}{width=0.8 \linewidth}
    \begin{tabular}{lllllccccccccccc}
    \toprule
    $\sigma$ &  Method  & Avg & Time (min) & MAD & ACR & 0.00  & 0.50 & 1.00 & 1.50 & 2.00 & 2.5 & 3.0 & 3.5 & 4.0\\ 
    \midrule
    \multirow{8}{*}{0.50}
  & $\rm{ISS}_{AD-10-0.05}$ & \textbf{32992}   & \textbf{317} & \textbf{0.05}   & \textbf{0.794} & 54.6 & 49.8 & 42.4 & \textbf{33.2} & 0.0 & 0.0 & 0.0 & 0.0 & 0.0\\
& $\rm{IAS}-{3.3}$ & 33000   & \textbf{317} & 0.14   & 0.77 & \textbf{54.8} & \textbf{50.2} & \textbf{43.4} & 32.8 & 0.0 & 0.0 & 0.0 & 0.0 & 0.0\\
\cmidrule(l){2-2} \cmidrule(l){3-15}
& $\rm{ISS}_{AD-10-0.10}$ & \textbf{22144}   & \textbf{213} & \textbf{0.10}   & \textbf{0.775} & 54.6 & 49.8 & 42.0 & \textbf{33.0} & 0.0 & 0.0 & 0.0 & 0.0 & 0.0\\
& $\rm{IAS}-{2.2}$ & 22200   & \textbf{213} & 0.19   & 0.752 & \textbf{54.8} & \textbf{50.2} & \textbf{43.4} & 32.6 & 0.0 & 0.0 & 0.0 & 0.0 & 0.0\\
\cmidrule(l){2-2} \cmidrule(l){3-15}
& $\rm{ISS}_{AD-50-0.05}$ & \textbf{144220}   & \textbf{1385} & \textbf{0.05}   & \textbf{0.856} & \textbf{54.8} & \textbf{50.2} & 42.8 & \textbf{34.2} & \textbf{29.8} & 0.0 & 0.0 & 0.0 & 0.0\\
& $\rm{IAS}-{14.4}$ & 144400   & 1386   & 0.15   & 0.831 & \textbf{54.8} & \textbf{50.2} & \textbf{43.4} & 33.4 & 0.0 & 0.0 & 0.0 & 0.0 & 0.0\\
\cmidrule(l){2-2} \cmidrule(l){3-15}
& $\rm{ISS}_{AD-50-0.10}$ & \textbf{95381}   & \textbf{916} & \textbf{0.10}   & \textbf{0.839} & \textbf{54.8} & \textbf{50.2} & 42.8 & \textbf{34.2} & 0.0 & 0.0 & 0.0 & 0.0 & 0.0\\
& $\rm{IAS}-{9.5}$ & 95400   & \textbf{916} & 0.20   & 0.815 & \textbf{54.8} & \textbf{50.2} & \textbf{43.4} & 33.2 & 0.0 & 0.0 & 0.0 & 0.0 & 0.0\\
    \midrule
    \multirow{8}{*}{1.00}
  & $\rm{ISS}_{AD-10-0.05}$ & \textbf{25987}   & \textbf{250} & \textbf{0.05}   & \textbf{0.958} & 40.6 & 36.8 & 31.8 & \textbf{28.0} & \textbf{24.0} & \textbf{20.2} & \textbf{17.4} & \textbf{13.4} & 0.0\\
& $\rm{IAS}-{2.6}$ & 26000   & \textbf{250} & 0.35   & 0.917 & \textbf{41.0} & \textbf{37.2} & \textbf{32.6} & \textbf{28.0} & \textbf{24.0} & 20.0 & 16.6 & 0.0 & 0.0\\
\cmidrule(l){2-2} \cmidrule(l){3-15}
& $\rm{ISS}_{AD-10-0.10}$ & \textbf{19209}   & \textbf{185} & \textbf{0.10}   & \textbf{0.943} & 40.2 & 36.8 & 31.8 & 27.4 & \textbf{24.0} & \textbf{20.0} & \textbf{17.4} & \textbf{13.4} & 0.0\\
& $\rm{IAS}-{1.9}$ & 19400   & 186   & 0.43   & 0.903 & \textbf{41.0} & \textbf{37.0} & \textbf{32.2} & \textbf{28.0} & \textbf{24.0} & \textbf{20.0} & 16.2 & 0.0 & 0.0\\
\cmidrule(l){2-2} \cmidrule(l){3-15}
& $\rm{ISS}_{AD-50-0.05}$ & \textbf{104037}   & \textbf{999} & \textbf{0.05}   & \textbf{1.015} & 40.6 & 36.8 & 32.2 & \textbf{28.0} & \textbf{24.0} & \textbf{20.4} & \textbf{17.4} & \textbf{13.4} & \textbf{13.4}\\
& $\rm{IAS}-{10.4}$ & 104200   & 1000   & 0.37   & 0.976 & \textbf{41.4} & \textbf{37.4} & \textbf{32.6} & \textbf{28.0} & \textbf{24.0} & \textbf{20.4} & \textbf{17.4} & \textbf{13.4} & 0.0\\
\cmidrule(l){2-2} \cmidrule(l){3-15}
& $\rm{ISS}_{AD-50-0.10}$ & \textbf{82899}   & \textbf{796} & \textbf{0.10}   & \textbf{1.005} & 40.6 & 36.8 & 32.2 & \textbf{28.0} & \textbf{24.0} & 20.0 & \textbf{17.4} & \textbf{13.4} & \textbf{13.4}\\
& $\rm{IAS}-{8.3}$ & 83000   & 797   & 0.43   & 0.967 & \textbf{41.4} & \textbf{37.4} & \textbf{32.6} & \textbf{28.0} & \textbf{24.0} & \textbf{20.4} & \textbf{17.4} & \textbf{13.4} & 0.0\\
    \bottomrule
    \end{tabular}
    \end{adjustbox}
    \vspace{-5pt}
\end{table*}

\subsection{Overall analysis of ACR and runtime}
Fig. \ref{fig:cifar_5}, Fig. \ref{fig:cifar_10}, Fig. \ref{fig:img_5}, Fig. \ref{fig:img_10} present the overall empirical results of ISS and IAS on CIFAR-10 and ImageNet.  As presented, ISS significantly accelerates the certification for randomized smoothing. Specifically, on ImageNet ($\sigma=0.5, 1.0$), $\rm{ISS}-10-0.05, \rm{ISS}-10-0.1$ reduce the original time cost $962$ minutes (the green dotted lines) to roughly $300, 200$ respectively at $\rm{U}_{AD}= 0.05, 0.1$ respectively. Overall, the speedups of ISS are even higher on CIFAR-10. We also compare ISS to IAS on two datasets. We found that ISS always achieves higher ACR than IAS in the similar time cost. For ImageNet ($\sigma=1.0$), $\rm{ISS}-20-0.05$ even further improves $\rm{IAS}-20$ by a moderate margin, while the time cost of $\rm{ISS}-20-0.05$ is only $0.56 \times$ of $\rm{IAS}-20$. The full results are reported in the supplemental material.


\subsection{Results of ISS ($\rm{AD}$) on ImageNet}
Table \ref{tab:imagenet_ard} reports the results of ISS\footnote{Here we only report the results at $\sigma=0.5$, and $\sigma=1.0$ because the work \cite{DBLP:conf/nips/JeongS20} only releases the training hyperparameters at $\sigma=0.5, 1.0$ for consistency training algorithm.}. Remarkably, ISS reduces the average sample size to roughly $\frac{3}{10} \times, \frac{1}{5} \times$ at the cost of $\rm{U}_{AD}=0.05, 0.10$ respectively, meaning the speedups are roughly $\frac{10}{3}\times, 5\times$. We found that the MADs of IAS are higher than ISS, meaning that IAS will cause a large radius decline on the specific inputs. Namely, the MAD of $\rm{IAS}-10.4$ is more than $7 \times$ $\rm{ISS}_{\rm{AD}-50-0.05}$. ISS consistently surpasses IAS on ACR. $\rm{ISS}_{\rm{AD}-50-0.10}(\sigma=1.0)$ achieves $\rm{ACR}=1.005$ in $796$ minutes while IAS only achieves $\rm{ACR}=0.976$ in $1,000$ minutes. We also observe that ISS slightly lower than IAS on the low-radius certified accuracies. It is because ISS tends to assign the small sample sizes to those inputs with low $p_A$, which inevitably sacrifices the certified radii of low-$p_A$ inputs. Meanwhile, ISS significantly improves the high-radius certified accuracies and ACR in return. 

\begin{table}[t!]
\centering
\caption{Experiment setting.}
    \vspace{-10pt}
\label{tab:config}
    \begin{adjustbox}{width=0.6\linewidth}
    \begin{tabular}{ l | c | c }
    \toprule
     Dataset & CIFAR-10 & ImageNet\\
     \hline
     Model & ResNe-110 & ResNet-50\\
     \hline
     Training by & MACER & Consistency\\
    \hline
    $\overline{k}$ & \multicolumn{2}{c}{100,000,\; 500,000}\\
    \hline
    $k_0$ &  \multicolumn{2}{c}{$0.01 \overline{k}$}\\
    \hline
    $\sigma$ & {0.25, 0.5, 1.0} & 0.5, 1.0\\
    \bottomrule
    \end{tabular}
    \end{adjustbox}
    \vspace{-5pt}
\end{table}
\begin{table}[t!]
\centering
\caption{Runtime for computing $\psi_{ISS}$.}
    \vspace{-10pt}
\label{tab:time}
    \begin{adjustbox}{width=0.5\linewidth}
    \begin{tabular}{ l | l | l | l}
    \toprule
      $\rm{AD}$ & Time (s) & $\rm{RD}$ & Time (s) \\
     \hline
    0.01 & 0.70 & 0.01 & 39.47\\
    0.05 & 0.65 & 0.05 & 13.52\\
    0.10 & 0.57 & 0.10 & 7.50\\
    \bottomrule
    \end{tabular}
    \end{adjustbox}
    \vspace{-10pt}
\end{table}

\begin{table*}[t!]
\centering
\caption{ImageNet: comparison on average sample size (Avg), certification runtime (in minutes), average certified radii (ACR) and certified accuracies ($\%$) on models trained by Consistency \cite{DBLP:conf/nips/JeongS20}. }
\label{tab:imagenet_rrd}
\vspace{-10pt}
    \begin{adjustbox}{width=0.8\linewidth}
    \begin{tabular}{lllllccccccccccccc}
    \toprule
    $\sigma$ &  Method  & Avg & Time & ACR & 0.00  & 0.50 & 1.00 & 1.50 & 2.00 & 2.5 & 3.0 & 3.5 & 4.0 \\ 
    \midrule
    \multirow{4}{*}{0.50}
   & $\rm{ISS}_{RD-10-0.01}$ & \textbf{70919}   & \textbf{682}  & \textbf{0.809} & \textbf{54.8} & \textbf{50.2} & \textbf{43.4} & \textbf{33.2} & 0.0 & 0.0 & 0.0 & 0.0 & 0.0\\
& $\rm{IAS}-{7.1}$ & 71000   & \textbf{682} & 0.803 & \textbf{54.8} & \textbf{50.2} & \textbf{43.4} & \textbf{33.2} & 0.0 & 0.0 & 0.0 & 0.0 & 0.0\\
\cmidrule(l){2-2} \cmidrule(l){3-14}
& $\rm{ISS}_{RD-10-0.05}$ & \textbf{34591}   & 333   & \textbf{0.781} & \textbf{54.8} & \textbf{50.2} & 42.2 & \textbf{33.0} & 0.0 & 0.0 & 0.0 & 0.0 & 0.0\\
& $\rm{IAS}-{3.5}$ & 34600   & \textbf{332} & 0.772 & \textbf{54.8} & \textbf{50.2} & \textbf{43.4} & 32.8 & 0.0 & 0.0 & 0.0 & 0.0 & 0.0\\
\midrule
\multirow{4}{*}{1.00}
& $\rm{ISS}_{RD-10-0.01}$ & \textbf{67207}   & \textbf{646}   & \textbf{0.966} & \textbf{41.4} & \textbf{37.4} & \textbf{32.6} & \textbf{28.0} & \textbf{24.0} & \textbf{20.4} & \textbf{17.4} & \textbf{13.4} & 0.0\\
& $\rm{IAS}-{6.7}$ & 67400   & 647   & 0.959 & 41.2 & \textbf{37.4} & \textbf{32.6} & \textbf{28.0} & \textbf{24.0} & \textbf{20.4} & \textbf{17.4} & \textbf{13.4} & 0.0\\
\cmidrule(l){2-2} \cmidrule(l){3-14}
& $\rm{ISS}_{RD-10-0.05}$ & \textbf{32515}   & \textbf{313} & \textbf{0.935} & \textbf{41.2} & 37.0 & 31.8 & 27.6 & 23.8 & \textbf{20.0} & \textbf{17.2} & \textbf{13.4} & 0.0\\
& $\rm{IAS}-{3.3}$ & 32600   & \textbf{313}  & 0.927 & 41.0 & \textbf{37.2} & \textbf{32.6} & \textbf{28.0} & \textbf{24.0} & \textbf{20.0} & 16.8 & \textbf{13.4} & 0.0\\
    \bottomrule
    \end{tabular}
    \end{adjustbox}
    \vspace{-10pt}
\end{table*}

\subsection{Results of ISS ($\rm{RD}$) on ImageNet}
Table \ref{tab:imagenet_rrd} reports the results of ISS ($\rm{RD}$) on ImageNet at $\rm{U}_{RD}=0.05, 0.10$. ISS reduces the average sample size to roughly $\frac{7}{10} \times, \frac{7}{20} \times$ at controllable cost of $\rm{RD}=1\%, 5\%$ respectively. Compared to IAS, ISS ($\rm{RD}$) also improves ACR. 

\begin{table*}[tb!]
\centering
\caption{CIFAR-10: comparison on average sample size (Avg), certification runtime (in minutes), maximum absolute decline (MAD), average certified radii (ACR) and certified accuracies ($\%$) on the models trained by MACER \cite{DBLP:conf/iclr/ZhaiDHZGRH020}. Results are evaluated on 500 testing data of $id=0, 20, \dots, 9960, 9980$. The bold denotes better performance.} 
\vspace{-10pt}
\label{tab:cifar_ard}
    \begin{adjustbox}{width=0.98\linewidth}
    \begin{tabular}{cllllcccccccccccccccccccc}
    \toprule
    $\sigma$ &  Method  & Avg & Time & MAD & ACR & 0.00 & 0.25 & 0.50 & 0.75 & 1.00 & 1.25 & 1.50 & 1.75 & 2.00 & 2.25 & 2.50 & 2.75 & 3.0 & 3.25 & 3.5 & 3.75 & 4.00\\ 
   \midrule
   \multirow{8}{*}{0.25}
    & $\rm{ISS}_{AD-10-0.05}$ & \textbf{22237}   & \textbf{39} & \textbf{0.05}   & \textbf{0.492} & 76.8 & 68.0 & 49.4 & \textbf{38.8} & 0.0 & 0.0 & 0.0 & 0.0 & 0.0 & 0.0 & 0.0 & 0.0 & 0.0 & 0.0 & 0.0 & 0.0 & 0.0\\
& $\rm{IAS}-{2.2}$ & 22400   & \textbf{39} & 0.10   & 0.483 & \textbf{77.8} & \textbf{68.6} & \textbf{52.0} & 37.6 & 0.0 & 0.0 & 0.0 & 0.0 & 0.0 & 0.0 & 0.0 & 0.0 & 0.0 & 0.0 & 0.0 & 0.0 & 0.0\\
\cmidrule(l){2-2} \cmidrule(l){3-23}
& $\rm{ISS}_{AD-10-0.10}$ & \textbf{10945}   & \textbf{19} & \textbf{0.10}   & \textbf{0.473} & 76.8 & 68.0 & 48.8 & \textbf{37.6} & 0.0 & 0.0 & 0.0 & 0.0 & 0.0 & 0.0 & 0.0 & 0.0 & 0.0 & 0.0 & 0.0 & 0.0 & 0.0\\
& $\rm{IAS}-{1.1}$ & 11000   & \textbf{19} & 0.15   & 0.462 & \textbf{77.4} & \textbf{68.4} & \textbf{51.6} & 36.8 & 0.0 & 0.0 & 0.0 & 0.0 & 0.0 & 0.0 & 0.0 & 0.0 & 0.0 & 0.0 & 0.0 & 0.0 & 0.0\\
\cmidrule(l){2-2} \cmidrule(l){3-23}
& $\rm{ISS}_{AD-50-0.05}$ & \textbf{98984}   & 172 & \textbf{0.05}   & \textbf{0.529} & 77.4 & 68.4 & 51.6 & \textbf{39.8} & 0.0 & 0.0 & 0.0 & 0.0 & 0.0 & 0.0 & 0.0 & 0.0 & 0.0 & 0.0 & 0.0 & 0.0 & 0.0\\
& $\rm{IAS}-{9.9}$ & 99000   & \textbf{171} & 0.10   & 0.518 & \textbf{77.8} & \textbf{69.0} & \textbf{52.2} & 39.4 & 0.0 & 0.0 & 0.0 & 0.0 & 0.0 & 0.0 & 0.0 & 0.0 & 0.0 & 0.0 & 0.0 & 0.0 & 0.0\\
\cmidrule(l){2-2} \cmidrule(l){3-23}
& $\rm{ISS}_{AD-50-0.10}$ & \textbf{46509}   & \textbf{81} & \textbf{0.10}   & \textbf{0.512} & 77.4 & 68.4 & 51.6 & \textbf{39.4} & 0.0 & 0.0 & 0.0 & 0.0 & 0.0 & 0.0 & 0.0 & 0.0 & 0.0 & 0.0 & 0.0 & 0.0 & 0.0\\
& $\rm{IAS}-{4.7}$ & 46600   & \textbf{81} & 0.14   & 0.501 & \textbf{77.8} & \textbf{68.8} & \textbf{52.0} & 38.8 & 0.0 & 0.0 & 0.0 & 0.0 & 0.0 & 0.0 & 0.0 & 0.0 & 0.0 & 0.0 & 0.0 & 0.0 & 0.0\\
    \midrule
    \multirow{8}{*}{0.50}
 & $\rm{ISS}_{AD-10-0.05}$ & \textbf{21836}   & \textbf{38} & \textbf{0.05}   & \textbf{0.647} & 60.6 & 53.0 & 46.8 & 39.8 & 32.4 & \textbf{26.0} & \textbf{19.8} & \textbf{13.0} & 0.0 & 0.0 & 0.0 & 0.0 & 0.0 & 0.0 & 0.0 & 0.0 & 0.0\\
& $\rm{IAS}-{2.2}$ & 22000   & \textbf{38} & 0.20   & 0.633 & \textbf{61.8} & \textbf{54.0} & \textbf{47.8} & \textbf{40.2} & \textbf{32.8} & \textbf{26.0} & 19.4 & 0.0 & 0.0 & 0.0 & 0.0 & 0.0 & 0.0 & 0.0 & 0.0 & 0.0 & 0.0\\
\cmidrule(l){2-2} \cmidrule(l){3-23}
& $\rm{ISS}_{AD-10-0.10}$ & \textbf{14620}   & \textbf{26} & \textbf{0.10}   & \textbf{0.634} & 60.6 & 53.0 & 46.8 & 39.4 & 31.0 & \textbf{26.0} & \textbf{19.6} & \textbf{11.2} & 0.0 & 0.0 & 0.0 & 0.0 & 0.0 & 0.0 & 0.0 & 0.0 & 0.0\\
& $\rm{IAS}-{1.5}$ & 14800   & \textbf{26} & 0.25   & 0.621 & \textbf{61.8} & \textbf{54.0} & \textbf{47.8} & \textbf{40.2} & \textbf{32.6} & \textbf{26.0} & 19.0 & 0.0 & 0.0 & 0.0 & 0.0 & 0.0 & 0.0 & 0.0 & 0.0 & 0.0 & 0.0\\
\cmidrule(l){2-2} \cmidrule(l){3-23}
& $\rm{ISS}_{AD-50-0.05}$ & \textbf{91293}   & \textbf{158} & \textbf{0.05}   & \textbf{0.68} & 61.8 & 54.0 & 47.6 & 40.0 & 32.6 & 26.0 & \textbf{20.2} & \textbf{14.2} & \textbf{10.2} & 0.0 & 0.0 & 0.0 & 0.0 & 0.0 & 0.0 & 0.0 & 0.0\\
& $\rm{IAS}-{9.1}$ & 91400   & \textbf{158} & 0.20   & 0.667 & \textbf{62.2} & \textbf{54.4} & \textbf{48.0} & \textbf{40.2} & \textbf{33.0} & \textbf{26.6} & 19.8 & 13.4 & 0.0 & 0.0 & 0.0 & 0.0 & 0.0 & 0.0 & 0.0 & 0.0 & 0.0\\
\cmidrule(l){2-2} \cmidrule(l){3-23}
& $\rm{ISS}_{AD-50-0.10}$ & \textbf{61567}   & \textbf{107} & \textbf{0.10}   & \textbf{0.673} & 61.8 & 54.0 & 47.6 & 40.0 & 32.6 & 25.2 & \textbf{20.0} & \textbf{13.8} & 0.0 & 0.0 & 0.0 & 0.0 & 0.0 & 0.0 & 0.0 & 0.0 & 0.0\\
& $\rm{IAS}-{6.2}$ & 61600   & \textbf{107} & 0.25   & 0.659 & \textbf{62.2} & \textbf{54.4} & \textbf{47.8} & \textbf{40.2} & \textbf{33.0} & \textbf{26.4} & 19.8 & 12.4 & 0.0 & 0.0 & 0.0 & 0.0 & 0.0 & 0.0 & 0.0 & 0.0 & 0.0\\
    \midrule
    \multirow{8}{*}{1.00}
 & $\rm{ISS}_{AD-10-0.05}$ & \textbf{21153}   & \textbf{37} & \textbf{0.05}   & \textbf{0.78} & 42.6 & 40.2 & 37.2 & 33.4 & 30.4 & 27.0 & 24.4 & 21.2 & \textbf{18.4} & \textbf{14.6} & \textbf{13.4} & \textbf{10.4} & \textbf{8.8} & \textbf{6.2} & \textbf{4.0} & \textbf{3.2} & 0.0\\
& $\rm{IAS}-{2.1}$ & 21200   & \textbf{37} & 0.40   & 0.763 & \textbf{42.8} & \textbf{40.6} & \textbf{37.4} & \textbf{34.0} & \textbf{31.0} & \textbf{27.4} & \textbf{24.8} & \textbf{21.4} & \textbf{18.4} & \textbf{14.6} & 12.8 & 9.8 & 8.0 & 4.6 & 0.0 & 0.0 & 0.0\\
\cmidrule(l){2-2} \cmidrule(l){3-23}
& $\rm{ISS}_{AD-10-0.10}$ & \textbf{14751}   & \textbf{26} & \textbf{0.10}   & \textbf{0.766} & 42.4 & 39.6 & 36.8 & 32.8 & 30.0 & 26.4 & 24.0 & 20.6 & 18.2 & \textbf{14.4} & \textbf{12.8} & \textbf{10.2} & \textbf{8.8} & \textbf{6.0} & \textbf{4.0} & 0.0 & 0.0\\
& $\rm{IAS}-{1.5}$ & 14800   & \textbf{26} & 0.50   & 0.754 & \textbf{42.8} & \textbf{40.4} & \textbf{37.4} & \textbf{33.8} & \textbf{30.8} & \textbf{27.4} & \textbf{24.8} & \textbf{21.4} & \textbf{18.4} & \textbf{14.4} & \textbf{12.8} & 9.6 & 7.2 & 3.8 & 0.0 & 0.0 & 0.0\\
\cmidrule(l){2-2} \cmidrule(l){3-23}
& $\rm{ISS}_{AD-50-0.05}$ & \textbf{70204}   & \textbf{123} & \textbf{0.05}   & \textbf{0.803} & \textbf{42.8} & 40.4 & \textbf{37.4} & 33.8 & 30.4 & 27.4 & 24.6 & \textbf{21.4} & \textbf{18.4} & 14.6 & 13.4 & \textbf{10.4} & \textbf{9.2} & \textbf{6.8} & \textbf{4.8} & \textbf{4.0} & \textbf{3.2}\\
& $\rm{IAS}-{7.0}$ & 70400   & \textbf{123} & 0.47   & 0.79 & \textbf{42.8} & \textbf{40.6} & \textbf{37.4} & \textbf{34.4} & \textbf{31.0} & \textbf{27.6} & \textbf{25.0} & \textbf{21.4} & \textbf{18.4} & \textbf{14.8} & \textbf{13.6} & 10.2 & 8.8 & 6.0 & 4.0 & 0.0 & 0.0\\
\cmidrule(l){2-2} \cmidrule(l){3-23}
& $\rm{ISS}_{AD-50-0.10}$ & \textbf{54102}   & \textbf{94} & \textbf{0.10}   & \textbf{0.797} & \textbf{42.8} & 40.4 & \textbf{37.4} & 33.8 & 30.4 & 27.4 & 24.6 & 21.2 & \textbf{18.4} & 14.4 & 13.0 & \textbf{10.0} & \textbf{9.2} & \textbf{6.6} & \textbf{4.8} & \textbf{4.0} & \textbf{3.2}\\
& $\rm{IAS}-{5.4}$ & 54200   & \textbf{94} & 0.53   & 0.785 & \textbf{42.8} & \textbf{40.6} & \textbf{37.4} & \textbf{34.4} & \textbf{31.0} & \textbf{27.6} & \textbf{25.0} & \textbf{21.4} & \textbf{18.4} & \textbf{14.8} & \textbf{13.4} & \textbf{10.0} & 8.6 & 5.6 & 4.0 & 0.0 & 0.0\\
    \bottomrule
    \end{tabular}
    \end{adjustbox}
    \vspace{-5pt}
\end{table*}

\subsection{Results of ISS ($\rm{AD}$) on CIFAR-10}
Table \ref{tab:cifar_ard} reports the results of ISS (OF $\rm{AD}$) on CIFAR-10. ISS reduces the average sample size to roughly $\frac{1}{5} \times, \frac{1}{10} \times$ at $\rm{U}_{AD}=0.05, 0.10$. Remarkably, ISS still improves ACRs and MADs, high-radius certified accuraries by a moderate margin on CIFAR-10. These empirical comparisons suggest that ISS is a better acceleration.


\subsection{Ablation study}

\paragraph{Choice on $\rm{AD}$ or $\rm{RD}$}
As shown in Fig. \ref{fig:ablation}, when $p_A: p_A \in [0.5, 1.0]$ increases, the sample size of ISS ($\rm{AD}$) monotonically increases, while the sample size of ISS ($\rm{RD}$) first decreases and then increases around $p_A=1.0$. ISS ($\rm{AD}$) can greatly improve ACR, but tends to sacrifice the certified radii of low-$p_A$ inputs a relatively larger proportion. ISS ($\rm{RD}$) sacrifices all inputs the same proportion of radius. 


\paragraph{Impact of $p_A$ and $\overline{k}$}
We investigate the impact of $p_A$ and $\overline{k}$ in Fig. \ref{fig:ablation} (\textbf{Upper}). For both $\rm{AD}$ and $\rm{RD}$, the sample size is $0$ when $p_A \leq 0.5$. It is because the certified radius is $0$ when $p_A \leq 0.5$. As expected, the sample size monotonically increases with $\overline{k}$ and decreases with $\rm{AD}$ (or $\rm{RD}$). 

\paragraph{Impact of $k_0/\overline{k}$}
We investigate the impact $k_0/\overline{k}$ on the runtime and ACR in Fig. \ref{fig:ablation} (\textbf{Lower}). Too small $k_0/\overline{k}$ results in a loose confidence interval $[p_{\rm{low}},p_{\rm{up}}]$, which can cause the ISS sample size $\hat{k}$ to be much larger than required. Too large $k_0/\overline{k}$ may waste too much computation in estimating $[p_{\rm{low}},p_{\rm{up}}]$. Our choice $k_0/\overline{k}=0.01$ performs well across various noise levels on CIFAR-10 and ImageNet.



\section{Conclusion}
\label{sec:conclusion}
Randomized smoothing has been suffering from the long certification runtime, but the current acceleration methods are low-efficiency. Therefore, we propose input-specific sampling, which adaptively assigns the sample size. Our work establishes an initial step towards a better performance-time trade-off for the certification of randomized smoothing. Specifically. Our strong empirical results suggest that ISS is a promising acceleratio. Specifically, ISS speeds up the certification by more than $4 \times$ only at the controllable cost of $0.10$ certified radius on ImageNet. An interesting direction for future work is to make the confidence interval estimation method adapt to the input.
\section*{Acknowledgements}
This work was supported by the National Key Research and Development Program of China No. 2020YFB1806700, Shanghai Municipal Science and Technology Major Project Grant 2021SHZDZX0102, NSFC Grant 61932014, NSFC Grant 61872241, Project BE2020026, the Key R$\&$D Program of Jiangsu, China. This work is also partially supported by The Hong Kong Polytechnic University under Grant P0030419, P0030929, and P0035358.
\bibliography{ref}
\bibstyle{aaai22}
\newpage
\end{document}